\pdfoutput=1
\documentclass[11pt]{article}

\usepackage[preprint]{acl}

\usepackage{times}
\usepackage{latexsym}
\usepackage{booktabs}
\usepackage{appendix}
\usepackage{amsmath}
\usepackage{tcolorbox}
\usepackage{fvextra}

\usepackage[T1]{fontenc}
\usepackage[utf8]{inputenc}

\usepackage{microtype}

\usepackage{inconsolata}

\usepackage{graphicx}

\title{Exploring Design of Multi-Agent LLM Dialogues for Research Ideation%
  \thanks{Accepted to SIGDIAL~2025}}
  
\author{
  \textbf{Keisuke Ueda\textsuperscript{1, 2}}
  \textbf{Wataru Hirota\textsuperscript{3}}
  \textbf{Takuto Asakura\textsuperscript{3}}
  \textbf{Takahiro Omi\textsuperscript{3}}
  \\
  \textbf{Kosuke Takahashi\textsuperscript{3}}
  \textbf{Kosuke Arima\textsuperscript{3}}
  \textbf{Tatsuya Ishigaki\textsuperscript{1}}
  \\
  \\
  \textsuperscript{1}Artificial Intelligence Research Center, AIST,
  \textsuperscript{2}EPFL,
  \textsuperscript{3}Stockmark,
  \\
  \small{keisuke.ueda@epfl.ch,}
  \small{ishigaki.tatsuya@aist.go.jp,}
  \\
  \small{\{wataru.hirota, takuto.asakura, takahiro.omi, kosuke.takahashi, kosuke.arima\}@stockmark.co.jp}
}

\begin{document}
\maketitle

\begin{abstract}
Large language models (LLMs) are increasingly used to support creative tasks such as research idea generation. While recent work has shown that structured dialogues between LLMs can improve the novelty and feasibility of generated ideas, the optimal design of such interactions remains unclear. In this study, we conduct a comprehensive analysis of multi-agent LLM dialogues for scientific ideation. We compare different configurations of agent roles, number of agents, and dialogue depth to understand how these factors influence the novelty and feasibility of generated ideas. Our experimental setup includes settings where one agent generates ideas and another critiques them, enabling iterative improvement. Our results show that enlarging the agent cohort, deepening the interaction depth, and broadening agent persona heterogeneity each enrich the diversity of generated ideas.  Moreover, specifically increasing critic-side diversity within the ideation–critique–revision loop further boosts the feasibility of the final proposals. Our findings offer practical guidelines for building effective multi-agent LLM systems for scientific ideation.
Our code is available at~\footnote{\url{https://github.com/g6000/MultiAgent-Research-Ideator}}.
\end{abstract}

\section{Introduction}

Large language models (LLMs) have recently demonstrated strong capabilities in supporting early-stage scientific ideation, such as proposing novel research directions or hypotheses based on seed topics~\cite{si2025can, wang2024scimon}. However, most prior work treats LLMs as single-shot generators, prompting a single model once to produce ideas without further refinement or critique. This overlooks the potential benefits of multi-agent collaboration, which is central to human creativity.

Recent studies have begun exploring interaction among LLMs through debate, self-critique, and role-play~\cite{li2023camel, du2023llmdebate, park2023generative}, showing promising results in tasks such as factual question answering and strategic planning. Despite these advances, it remains unclear how to best structure such interactions for open-ended creative tasks like research ideation.

In this work, we conduct a controlled study to investigate how multi-agent LLM dialogues should be designed to improve the quality of generated research ideas. We systematically explore the impact of three key design dimensions by varying the dialogue structure, corresponding to our main experimental manipulations:

\begin{itemize}
    \item \textbf{Agent Diversity}: We study how injecting domain-specific personas (e.g., ``Physics-AI'', ``Psychology-AI'') into the dialogue affects ideation quality.
    \item \textbf{Agent Parallelism}: We vary the number of simultaneous critique agents that independently review the initial ideas before a single revision step, analyzing how the breadth of the concurrent feedback shapes the outcome.
    \item \textbf{Agent Interaction Depth}: We change the number of critique–revision turns executed in sequence, thereby controlling how deeply the ideas are iteratively refined by the same set of agents.
\end{itemize}

While these design aspects—diversity, parallelism/breadth, and depth—are frequently configured heuristically in prior multi-agent systems, their individual contributions and potential interactions in the context of open-ended creative generation like research ideation remain largely unexplored. Our study aims to disentangle these effects through controlled comparisons across various configurations within a structured dialogue framework, providing empirical guidance on effective dialogue design.

\begin{figure}
    \centering
    \includegraphics[width=0.95 \linewidth]{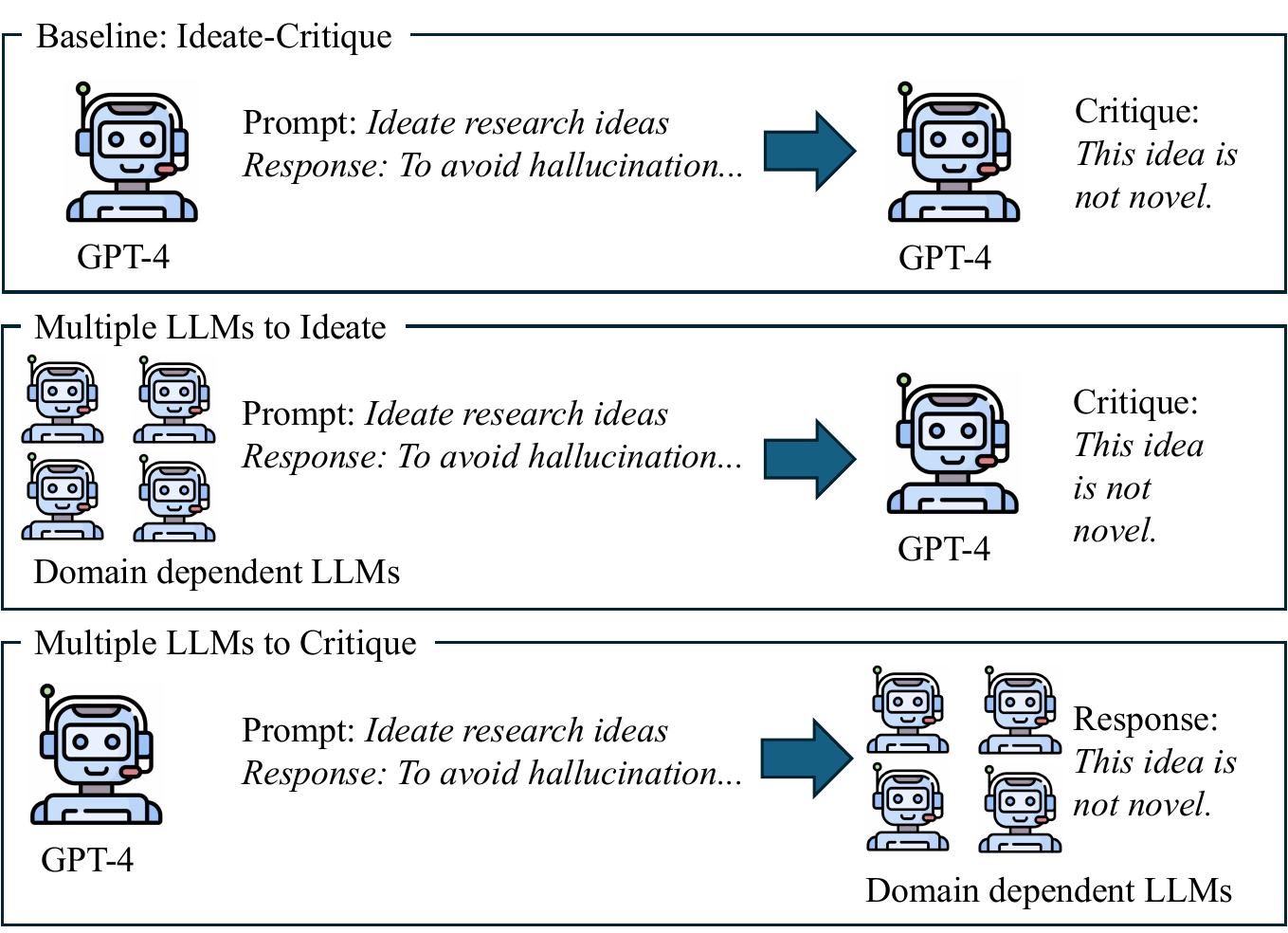}
    \caption{Examples of LLMs' discussion-based ideation. We compare several combinations of the number of LLMs, their assigned personas (e.g., domain-generalist or domain-expert), and the number of times that the \textit{ideation–critique–revision} cycle is repeated}
    \label{fig:sample-discussions}
\end{figure}

To address this gap, we adopt a structured \textit{ideation–critique–revision} framework and evaluate seven dialogue configurations across seven diverse research topics in AI and NLP. Each configuration varies in agent count, persona specialization, and dialogue depth. Outputs are evaluated automatically using diversity metrics and a GPT-4 preference tournament.

Our goal is not to propose a novel generation framework, but rather to provide empirical guidance for designing effective multi-agent LLM ideation systems. Our findings show that involving multiple agents with complementary domain expertise and allowing for multi-turn refinement significantly improves the quality of generated ideas. These results offer practical insights for the development of LLM-based tools to support scientific creativity.

\section{Related Work}

\paragraph{AI-driven scientific discovery.}
From the robot scientist \emph{Adam}~\cite{king2009automation} to calls for an ``AI third pillar of science''~\cite{gil2014amplify,xu2021powerful,kitano2021nobelturing}, work on automating discovery has accelerated.
Surveys map current progress and challenges across disciplines~\cite{wang2023scientific,eger2025transforming}.
LLMs already synthesize papers and data for molecular property prediction~\cite{zheng2023large}, solve scientist-curated coding tasks~\cite{tian2024scicode}, and beat humans in rapid ML-engineering settings~\cite{wijk2024rebench}.
Auxiliary tools such as GraphEval~\cite{feng2025grapheval}, SciMON~\cite{wang2024scimon}, and Self-Refine~\cite{madaan2023selfrefine} quantify or boost idea quality, yet they all rely on single-agent prompting.

\paragraph{Multi-agent interaction with LLMs.}
Classical MAS surveys and logics emphasize coordination, commitment and diversity~\cite{dorri2018survey,duninkeplicz2011teamwork}, echoing evidence that heterogeneous groups innovate better~\cite{page2008diversity,paulus2019creativity}.
For LLMs, debate~\cite{du2023llmdebate}, role-play societies (CAMEL)~\cite{li2023camel}, memory-rich generative agents~\cite{park2023generative}, and critical discussions~\cite{wang2024rethinking} raise accuracy or behavioral realism, but mostly on factual or planning tasks—not open-ended ideation.

\paragraph{Multi-agent automation of research.}
Full-pipeline systems chain specialized agents to write publishable papers (AI Scientist~\cite{lu2024aiscientist}, its successor AI Scientist-v2~\cite{yamada2025aiscientistv2}, and AI-Researcher~\cite{tang2025airesearcher}), iterate with an automated reviewer (CycleResearcher~\cite{weng2025cycleresearcher}), share results via a communal memory (agentRxiv~\cite{schmidgall2025agentrxiv}) or a review–experiment–write loop (Agent Laboratory~\cite{schmidgall2025agentlab}).
Community-level simulators model entire research graphs~\cite{yu2024researchtown}, while AI Co-Scientist validates biomedical hypotheses in vitro~\cite{gottweis2025aicoscientist}. 
VirSci (Many Heads Are Better Than One)~\cite{su2025manyheads} focuses specifically on scientific idea generation, letting diverse ``virtual scientists'' iteratively refine proposals and surpass single-agent baselines.
Despite these advances--and promising single-agent creativity studies~\cite{si2025can}--no work has systematically measured how agent \emph{diversity, parallelism}, and \emph{interaction depth} trade off when generating research ideas.
Our experiments close this gap.

%\textbf{AI Support for Scientific Creativity.} Research also explores integrating LLMs into tools that support scientific workflows. Gu and Krenn~\cite{gu2024scimuse} used knowledge graphs to guide LLMs in generating personalized research ideas, evaluated by human experts. Wang et al.~\cite{wang2019paperrobot} proposed a system that incrementally drafts scientific content from literature. Other systems, such as Galactica~\cite{taylor2022galactica}, aim to serve as interfaces for scientific knowledge retrieval and composition. Our work differs in focusing specifically on the design of autonomous AI-AI dialogues for early-stage research ideation.

\section{Model Setup}

\subsection{Overall Framework}

We adopt the research ideation setup introduced by~\citet{si2025can} as the basis for our experimental design. In their framework, a seed query or topic is used to retrieve a set of related papers via the Semantic Scholar API~\footnote{\url{https://www.semanticscholar.org/product/api}}. From these, $n$ papers (typically $n=10$) are randomly selected, and a large language model (LLM) is prompted to generate $m$ research ideas per the retrieved set of papers. This process is repeated until a total of $R$ ideas (e.g., 4,000) are collected.

Following their pipeline, we apply an embedding-based similarity filter to remove redundant or near-duplicate ideas. The remaining ideas are evaluated using an LLM-as-a-judge rubric to assign scores for originality, feasibility, and clarity. Top-ranked ideas are then expanded into full research proposal drafts.

This framework serves as a consistent foundation across all dialogue configurations we compare in this study. Our focus is not on proposing new pipeline but on analyzing how different multi-agent dialogue settings—when applied within this structure—affect the quality of generated ideas.

\subsection{LLMs' Discussion-based Ideation–Critique–Revision}
\label{sec:model}

Although effective, the original framework suffers from a high redundancy rate, with most generated ideas being discarded during deduplication as mentioned in the original literature~\cite{su2025manyheads}. This highlights a critical need to improve the quality and diversity of the initial ideation module. In the baseline approach by \citet{si2025can}, a single LLM is responsible for idea generation, which limits the potential for exploration and refinement.

To address this, we explore discussion-based ideation frameworks that replace the single-agent generation module with a multi-agent dialogue. Specifically, we adopt a structure involving \textit{Ideation–Critique–Revision}. In this framework, one or more LLMs first propose ideas, a second set of LLMs critiques them, and then the ideas are revised based on the critiques.

We systematically investigate the impact of different dialogue configurations by implementing and comparing the following variants:

\begin{itemize}
    \item \textbf{Single (No Critique)}: A single LLM generates an idea without critique. This replicates the method used in~\citet{si2025can}.
    
    \item \textbf{Baseline (Self-Critique)}: A single LLM performs ideation, then critiques its own ideas, and finally revises them based on its self-critique. This simulates a simple form of self-reflection.
    
    \item \textbf{Iterative Self-Critique ($L=2, 3, 4$)}: Corresponding to \textbf{Agent Interaction Depth}, the ideation–critique–revision cycle by a single LLM is repeated $L$ times ($L=2, 3, 4$), allowing for deeper iterative refinement.
    
    \item \textbf{Parallel Self-Critique ($N=2, 3, 4$)}: Corresponding to \textbf{Agent Parallelism}, $N$ independent critique processes ($N=2, 3, 4$) are performed in parallel by the same LLM after initial ideation. The critiques are aggregated before a single revision step, simulating broader feedback.
    
    \item \textbf{Diverse Personas}: Investigating \textbf{Agent Diversity}, we assign a specific domain persona (e.g., ``Physics-AI'') to either the \textit{critic} role or the \textit{proposer/reviser} role(s). The remaining role(s) are filled by a standard ``AI researcher'' agent. This allows comparing the impact of persona injection at different stages of the dialogue.
\end{itemize}

These configurations allow us to systematically study how \textbf{Agent Diversity} (through persona injection), \textbf{Agent Parallelism} (in critique), and \textbf{Agent Interaction Depth} (iterative refinement) influence the novelty and usefulness of generated research ideas within a structured dialogue framework.

\section{Experiments}

\subsection{Data and Settings}

To evaluate our discussion-based ideation framework, we follow the open-sourced pipeline described in Section~\ref{sec:model}. Following \citet{si2025can}, we consider the same seven research topics in AI and NLP (\textit{bias, coding, safety, multilinguality, factuality, math, uncertainty}). For each topic we instantiate all ten dialogue configurations listed in Section~\ref{sec:model}. 

For every \textit{topic} $\times$ \textit{configuration} pair we run 20 independent trials with different random seeds. Each trial generates $k=5$ candidate ideas using GPT-4o-mini with retrieval-augmented prompting (10 papers sampled from a Semantic Scholar paper bank). This yields $7 \text{ topics} \times 10 \text{ configs} \times 20 \text{ seeds} \times 5 = 7{,}000$ raw ideas in total.

After generation, we apply the two-stage post-processing described earlier: (i) semantic deduplication (cosine similarity $\textgreater 0.8$ using MiniLM embeddings) and (ii) expansion of the remaining ideas into detailed experiment proposals. All proposals are produced with identical decoding parameters to ensure comparability.

\subsection{Evaluation}

We rely on two automatic evaluation protocols.

\paragraph{Diversity.} We report the \textit{Non-Duplicate Ratio}: the percentage of ideas that survive the embedding-based deduplication filter. A higher value indicates that a configuration produces more distinct ideas.

\paragraph{LLM Preference Ranking.} Quality is assessed via an LLM-as-a-judge tournament~\cite{si2025can}. For each topic we form A/B pairs between proposals from the \textbf{Baseline} configuration and every other configuration. An impartial GPT-4 model receives each pair and selects the better idea in a zero-shot setting; the procedure is repeated for ten elimination rounds with the preference score of each proposal accumulated across matches. This produces a scalar \textit{AI-ranking score} for every proposal and allows us to compute \textbf{Precision@N}: the fraction of the top-$N$ ranked proposals that come from the non-Baseline configuration (we report $N\!\in\!\{10,20,40\}$).

We report aggregate statistics (mean AI-ranking score and win rate against the Baseline configuration) for each configuration. No human annotation is involved in this study.

\section{Results}

Across all seven research topics, the three design axes introduced in Section~\ref{sec:model} exhibit clear and largely orthogonal effects on both diversity and quality. Below we summarize the main findings; all numbers refer to averages over topics.

\paragraph{Overall Trends.} The \textbf{Baseline} self-critique yields a solid improvement over the \textbf{Single} setting, increasing the Non-Duplicate Ratio by +0.08 (Table~\ref{tab:parallelism}) while also serving as a competitive reference point for quality. Every enhanced configuration further improves at least one evaluation dimension relative to Baseline, demonstrating that the proposed axes capture meaningful levers for idea generation.

\paragraph{Effect of Agent Parallelism.} Increasing the number of parallel critics consistently raises diversity: moving from one to three critics boosts the Non-Duplicate Ratio from 0.77 to 0.80, after which returns diminish Table~\ref{tab:parallelism}. The same trend is reflected in Precision@20, which rises from 0.47 (two critics, therefore marginally worse than Baseline) to the neutrality point of 0.50 (three critics) before dropping slightly at four critics. These results suggest that three independent viewpoints strike a good balance between breadth of feedback and noise.

\paragraph{Effect of Agent Interaction Depth.} Repeating the critique–revision loop improves both diversity and quality up to three iterations (Table~\ref{tab:depth}). At $L{=}3$ we observe a high Non-Duplicate Ratio (0.85) and the best Precision@10 (0.52). A fourth iteration offers negligible gains and sometimes hurts precision, indicating diminishing returns once the dialogue has converged.

\paragraph{Effect of Agent Diversity.} Injecting a specialized persona either at the critic stage or at the proposer/reviser stage affects diversity and precision differently (Table~\ref{tab:diversity}). A specialized critic delivers the strongest quality gains, achieving a win rate of 0.55 against Baseline at $N{=}10$, but offers little improvement in diversity. Conversely, assigning the persona to the proposer/reviser yields the highest diversity (0.81) while maintaining a competitive precision. These findings highlight a trade-off between novelty and focused quality feedback when choosing where to introduce domain expertise.

\paragraph{Summary.} Combining a moderate amount of parallel critique (three critics) with two to three refinement turns delivers the best overall performance, while persona diversity provides complementary gains in either diversity or precision depending on its placement in the dialogue. These results offer actionable guidance for practitioners seeking to deploy multi-agent LLM systems for creative ideation tasks.

\begin{table}[t]
\centering
\resizebox{\linewidth}{!}{
\begin{tabular}{lccc}
\toprule
N             & Non-Dup. Ratio & Precision@(10/20/40) \\
\midrule
Single  ($N{=}0$)       & 0.69          & - \\
Baseline ($N{=}1$)      & 0.77          & - \\
2             & 0.78          & 0.42 / 0.47 / 0.48 \\
3             & \textbf{0.80} & 0.41 / 0.47 / 0.50 \\
4             & 0.79          & 0.44 / 0.49 / 0.49 \\
\bottomrule
\end{tabular}
}
\caption{Impact of agent parallelism (number of critics) on diversity and Precision@N}
\label{tab:parallelism}
\end{table}

\begin{table}[t]
\centering
\resizebox{\linewidth}{!}{
\begin{tabular}{lccc}
\toprule
L (turns) & Non-Dup. Ratio & Precision@(10/20/40) \\
\midrule
Single ($L{=}0$)        & 0.69           & - \\
Baseline ($L{=}1$)     & 0.77           & - \\
2             & 0.80           & 0.49 / 0.48 / 0.50 \\
3             & 0.85           & \textbf{0.52} / 0.47 / 0.50 \\
4             & \textbf{0.86}  & 0.48 / 0.48 / 0.49 \\
\bottomrule
\end{tabular}
}
\caption{Impact of interaction depth (number of critique--revision turns) on metrics}
\label{tab:depth}
\end{table}

\begin{table}[t]
\centering
\resizebox{\linewidth}{!}{
\begin{tabular}{lccc}
\toprule
Configuration & Non-Dup. Ratio & Precision@(10/20/40) \\
\midrule
Baseline & 0.77 & - \\
Diverse Critic & 0.76  & \textbf{0.55} / \textbf{0.54} / \textbf{0.52} \\
Diverse Prop/Rev & \textbf{0.81} & 0.45 / 0.50 / 0.49 \\
\bottomrule
\end{tabular}
}
\caption{Impact of persona diversity on diversity and Precision@N}
\label{tab:diversity}
\end{table}

\section{Conclusion}

We presented a controlled, factorial study on how to design multi-agent LLM dialogues for scientific ideation. Our results show that enlarging the agent cohort, deepening the interaction depth, and broadening persona heterogeneity each enrich the diversity of generated ideas. Moreover, specifically increasing critic-side diversity within the ideation–critique–revision loop further boosts the feasibility of the final proposals. We provide empirical evidence that both agent diversity and dialogue depth contribute to more useful research ideas.

\section{Limitations and Future Directions}

Our study has at least two main limitations.

\paragraph{Automatic quality assessment.} Idea quality is evaluated exclusively via GPT-4 preference tournaments.  Although prior work reports moderate correlation between GPT-4 judgments and expert ratings, relying on a single automatic judge introduces model bias and potential circularity. Conducting a small-scale human evaluation remains an important next step.

\paragraph{Scope of dialogue configurations.} We explore three axes (diversity, parallelism, depth) at a limited set of levels chosen for tractability rather than to test a particular cognitive theory of creativity. Future work could ground the factor selection in formal models (e.g., collective intelligence or brainstorming literature) and study richer interactions such as argumentation or hierarchical planning.

\bibliography{custom}

\begin{thebibliography}{30}
\providecommand{\natexlab}[1]{#1}

\bibitem[{Dorri et~al.(2018)Dorri, Kanhere, and Jurdak}]{dorri2018survey}
Ali Dorri, Salil~S. Kanhere, and Raja Jurdak. 2018.
\newblock \href {https://doi.org/10.1109/ACCESS.2018.2831228} {Multi-agent systems: A survey}.
\newblock \emph{IEEE Access}, 6:28573--28593.

\bibitem[{Du et~al.(2023)Du, Li, Torralba, Tenenbaum, and Mordatch}]{du2023llmdebate}
Yilun Du, Shuang Li, Antonio Torralba, Joshua~B. Tenenbaum, and Igor Mordatch. 2023.
\newblock \href {https://arxiv.org/abs/2305.14325} {Improving factuality and reasoning in language models through multiagent debate}.
\newblock \emph{Preprint}, arXiv:2305.14325.

\bibitem[{Dunin-Keplicz and Verbrugge(2011)}]{duninkeplicz2011teamwork}
Barbara Dunin-Keplicz and Rineke Verbrugge. 2011.
\newblock \href {http://dx.doi.org/10.1002/9780470665237} {\emph{Teamwork in Multi-Agent Systems: A Formal Approach}}.
\newblock Wiley.

\bibitem[{Eger et~al.(2025)Eger, Cao, D'Souza, Geiger, Greisinger, Gross, Hou, Krenn, Lauscher, Li, Lin, Moosavi, Zhao, and Miller}]{eger2025transforming}
Steffen Eger, Yong Cao, Jennifer D'Souza, Andreas Geiger, Christian Greisinger, Stephanie Gross, Yufang Hou, Brigitte Krenn, Anne Lauscher, Yizhi Li, Chenghua Lin, Nafise~Sadat Moosavi, Wei Zhao, and Tristan Miller. 2025.
\newblock \href {https://arxiv.org/abs/2502.05151} {Transforming science with large language models: A survey on ai-assisted scientific discovery, experimentation, content generation, and evaluation}.
\newblock \emph{Preprint}, arXiv:2502.05151.

\bibitem[{Feng et~al.(2025)Feng, Sun, and You}]{feng2025grapheval}
Tao Feng, Yihang Sun, and Jiaxuan You. 2025.
\newblock \href {https://arxiv.org/abs/2503.12600} {Grapheval: A lightweight graph-based llm framework for idea evaluation}.
\newblock In \emph{The Thirteenth International Conference on Learning Representations}.

\bibitem[{Gil et~al.(2014)Gil, Greaves, Hendler, and Hirsh}]{gil2014amplify}
Yolanda Gil, Mark Greaves, James Hendler, and Haym Hirsh. 2014.
\newblock \href {https://doi.org/10.1126/science.1259439} {Amplify scientific discovery with artificial intelligence}.
\newblock \emph{Science}, 346(6206):171--172.

\bibitem[{Gottweis et~al.(2025)Gottweis, Weng, Daryin, Tu, Palepu, Sirkovic, Myaskovsky, Weissenberger, Rong, Tanno, Saab, Popovici, Blum, Zhang, Chou, Hassidim, Gokturk, Vahdat, Kohli, Matias, Carroll, Kulkarni, Tomasev, Guan, Dhillon, Vaishnav, Lee, Costa, Penadés, Peltz, Xu, Pawlosky, Karthikesalingam, and Natarajan}]{gottweis2025aicoscientist}
Juraj Gottweis, Wei-Hung Weng, Alexander Daryin, Tao Tu, Anil Palepu, Petar Sirkovic, Artiom Myaskovsky, Felix Weissenberger, Keran Rong, Ryutaro Tanno, Khaled Saab, Dan Popovici, Jacob Blum, Fan Zhang, Katherine Chou, Avinatan Hassidim, Burak Gokturk, Amin Vahdat, Pushmeet Kohli, and 15 others. 2025.
\newblock \href {https://arxiv.org/abs/2502.18864} {Towards an ai co-scientist}.
\newblock \emph{Preprint}, arXiv:2502.18864.

\bibitem[{King et~al.(2009)King, Rowland, Oliver, Young, Aubrey, Byrne, Liakata, Markham, Pir, Soldatova, Sparkes, Whelan, and Clare}]{king2009automation}
Ross~D. King, Jem Rowland, Stephen~G. Oliver, Michael Young, Wayne Aubrey, Emma Byrne, Maria Liakata, Magdalena Markham, Pinar Pir, Larisa~N. Soldatova, Andrew Sparkes, Kenneth~E. Whelan, and Amanda Clare. 2009.
\newblock \href {https://doi.org/10.1126/science.1165620} {The automation of science}.
\newblock \emph{Science}, 324(5923):85--89.

\bibitem[{Kitano(2021)}]{kitano2021nobelturing}
Hiroaki Kitano. 2021.
\newblock \href {https://doi.org/10.1038/s41540-021-00189-3} {The nobel turing challenge: Creating the engine for scientific discovery}.
\newblock \emph{npj Systems Biology and Applications}, 7:29.

\bibitem[{Li et~al.(2023)Li, Al~Kader~Hammoud, Itani, Khizbullin, and Ghanem}]{li2023camel}
Guohao Li, Hasan~Abed Al~Kader~Hammoud, Hani Itani, Dmitrii Khizbullin, and Bernard Ghanem. 2023.
\newblock \href {https://dl.acm.org/doi/10.5555/3666122.366838} {Camel: communicative agents for "mind" exploration of large language model society}.
\newblock In \emph{Proceedings of the 37th International Conference on Neural Information Processing Systems}, NeurIPS '23, Red Hook, NY, USA. Curran Associates Inc.

\bibitem[{Lu et~al.(2024)Lu, Lu, Lange, Foerster, Clune, and Ha}]{lu2024aiscientist}
Chris Lu, Cong Lu, Robert~Tjarko Lange, Jakob Foerster, Jeff Clune, and David Ha. 2024.
\newblock \href {https://arxiv.org/abs/2408.06292} {The ai scientist: Towards fully-automated open-ended scientific discovery}.

\bibitem[{Madaan et~al.(2023)Madaan, Tandon, Gupta, Hallinan, Gao, Wiegreffe, Alon, Dziri, Prabhumoye, Yang, Gupta, Majumder, Hermann, Welleck, Yazdanbakhsh, and Clark}]{madaan2023selfrefine}
Aman Madaan, Niket Tandon, Prakhar Gupta, Skyler Hallinan, Luyu Gao, Sarah Wiegreffe, Uri Alon, Nouha Dziri, Shrimai Prabhumoye, Yiming Yang, Shashank Gupta, Bodhisattwa~Prasad Majumder, Katherine Hermann, Sean Welleck, Amir Yazdanbakhsh, and Peter Clark. 2023.
\newblock \href {https://openreview.net/forum?id=S37hOerQLB} {Self-refine: Iterative refinement with self-feedback}.
\newblock In \emph{Thirty-seventh Conference on Neural Information Processing Systems}.

\bibitem[{Page(2008)}]{page2008diversity}
Scott~E. Page. 2008.
\newblock \href {https://doi.org/10.2307/j.ctt7sp9c} {\emph{The Difference: How the Power of Diversity Creates Better Groups, Firms, Schools, and Societies}}.
\newblock Princeton University Press.

\bibitem[{Park et~al.(2023)Park, O'Brien, Cai, Morris, Liang, and Bernstein}]{park2023generative}
Joon~Sung Park, Joseph~C. O'Brien, Carrie~J. Cai, Meredith~Ringel Morris, Percy Liang, and Michael~S. Bernstein. 2023.
\newblock \href {https://doi.org/10.1145/3586183.3606763} {Generative agents: Interactive simulacra of human behavior}.
\newblock In \emph{In the 36th Annual ACM Symposium on User Interface Software and Technology (UIST '23)}, UIST '23, New York, NY, USA. Association for Computing Machinery.

\bibitem[{Paulus and Nijstad(2019)}]{paulus2019creativity}
Paul~B. Paulus and Bernard~A. Nijstad. 2019.
\newblock \href {https://doi.org/10.1093/oxfordhb/9780190648077.001.0001} {\emph{The Oxford Handbook of Group Creativity and Innovation}}.
\newblock Oxford University Press.

\bibitem[{Schmidgall and Moor(2025)}]{schmidgall2025agentrxiv}
Samuel Schmidgall and Michael Moor. 2025.
\newblock \href {https://arxiv.org/abs/2503.18102} {Agentrxiv: Towards collaborative autonomous research}.
\newblock \emph{Preprint}, arXiv:2503.18102.

\bibitem[{Schmidgall et~al.(2025)Schmidgall, Su, Wang, Sun, Wu, Yu, Liu, Moor, Liu, and Barsoum}]{schmidgall2025agentlab}
Samuel Schmidgall, Yusheng Su, Ze~Wang, Ximeng Sun, Jialian Wu, Xiaodong Yu, Jiang Liu, Michael Moor, Zicheng Liu, and Emad Barsoum. 2025.
\newblock \href {https://arxiv.org/abs/2501.04227} {Agent laboratory: Using llm agents as research assistants}.
\newblock \emph{Preprint}, arXiv:2501.04227.

\bibitem[{Si et~al.(2025)Si, Yang, and Hashimoto}]{si2025can}
Chenglei Si, Diyi Yang, and Tatsunori Hashimoto. 2025.
\newblock \href {https://openreview.net/forum?id=M23dTGWCZy} {Can {LLM}s generate novel research ideas? a large-scale human study with 100+ {NLP} researchers}.
\newblock In \emph{The Thirteenth International Conference on Learning Representations}.

\bibitem[{Su et~al.(2025)Su, Chen, Tang, Yin, Zheng, Li, Qi, Wu, Li, Ouyang, Torr, Zhou, and Dong}]{su2025manyheads}
Haoyang Su, Renqi Chen, Shixiang Tang, Zhenfei Yin, Xinzhe Zheng, Jinzhe Li, Biqing Qi, Qi~Wu, Hui Li, Wanli Ouyang, Philip Torr, Bowen Zhou, and Nanqing Dong. 2025.
\newblock \href {https://arxiv.org/abs/2410.09403} {Many heads are better than one: Improved scientific idea generation by a llm-based multi-agent system}.
\newblock In \emph{Proceedings of the 63nd Annual Meeting of the Association for Computational Linguistics (Volume 1: Long Papers)}.

\bibitem[{Tang et~al.(2025)Tang, Xia, Li, and Huang}]{tang2025airesearcher}
Jiabin Tang, Lianghao Xia, Zhonghang Li, and Chao Huang. 2025.
\newblock \href {https://arxiv.org/abs/2505.18705} {{AI-Researcher: Autonomous Scientific Innovation}}.
\newblock \emph{Preprint}, arXiv:2505.18705.

\bibitem[{Tian et~al.(2024)Tian, Gao, Zhang, Chen, Fan, Guo, Haas, Ji, Krongchon, Li, Liu, Luo, Ma, Tong, Trinh, Tian, Wang, Wu, Xiong, Yin, Zhu, Lieret, Lu, Liu, Du, Tao, Press, Callan, Huerta, and Peng}]{tian2024scicode}
Minyang Tian, Luyu Gao, Shizhuo~Dylan Zhang, Xinan Chen, Cunwei Fan, Xuefei Guo, Roland Haas, Pan Ji, Kittithat Krongchon, Yao Li, Shengyan Liu, Di~Luo, Yutao Ma, Hao Tong, Kha Trinh, Chenyu Tian, Zihan Wang, Bohao Wu, Yanyu Xiong, and 11 others. 2024.
\newblock \href {https://arxiv.org/abs/2407.13168} {Scicode: A research coding benchmark curated by scientists}.
\newblock \emph{Preprint}, arXiv:2407.13168.

\bibitem[{Wang et~al.(2023)Wang, Fu, Du, Gao, Huang, Liu, Chandak, Liu, Van~Katwyk, Deac, Anandkumar, Bergen, Gomes, Ho, Kohli, Lasenby, Leskovec, Liu, Manrai, Marks, Ramsundar, Song, Sun, Tang, Veli{\v{c}}kovi{\'c}, Welling, Zhang, Coley, Bengio, and Zitnik}]{wang2023scientific}
Hanchen Wang, Tianfan Fu, Yuanqi Du, Wenhao Gao, Kexin Huang, Ziming Liu, Payal Chandak, Shengchao Liu, Peter Van~Katwyk, Andreea Deac, Anima Anandkumar, Karianne Bergen, Carla~P. Gomes, Shirley Ho, Pushmeet Kohli, Joan Lasenby, Jure Leskovec, Tie-Yan Liu, Arjun Manrai, and 11 others. 2023.
\newblock \href {https://doi.org/10.1038/s41586-023-06221-2} {Scientific discovery in the age of artificial intelligence}.
\newblock \emph{Nature}, 620(7972):47--60.

\bibitem[{Wang et~al.(2024{\natexlab{a}})Wang, Wang, Su, Tong, and Song}]{wang2024rethinking}
Qineng Wang, Zihao Wang, Ying Su, Hanghang Tong, and Yangqiu Song. 2024{\natexlab{a}}.
\newblock \href {https://doi.org/10.18653/v1/2024.acl-long.331} {Rethinking the bounds of {LLM} reasoning: Are multi-agent discussions the key?}
\newblock In \emph{Proceedings of the 62nd Annual Meeting of the Association for Computational Linguistics (Volume 1: Long Papers)}, pages 6106--6131, Bangkok, Thailand. Association for Computational Linguistics.

\bibitem[{Wang et~al.(2024{\natexlab{b}})Wang, Downey, Ji, and Hope}]{wang2024scimon}
Qingyun Wang, Doug Downey, Heng Ji, and Tom Hope. 2024{\natexlab{b}}.
\newblock \href {https://doi.org/10.18653/v1/2024.acl-long.18} {{S}ci{MON}: Scientific inspiration machines optimized for novelty}.
\newblock In \emph{Proceedings of the 62nd Annual Meeting of the Association for Computational Linguistics (Volume 1: Long Papers)}, pages 279--299, Bangkok, Thailand. Association for Computational Linguistics.

\bibitem[{Weng et~al.(2025)Weng, Zhu, Bao, Zhang, Wang, Zhang, and Yang}]{weng2025cycleresearcher}
Yixuan Weng, Minjun Zhu, Guangsheng Bao, Hongbo Zhang, Jindong Wang, Yue Zhang, and Linyi Yang. 2025.
\newblock \href {https://openreview.net/forum?id=bjcsVLoHYs} {Cycleresearcher: Improving automated research via automated review}.
\newblock In \emph{The Thirteenth International Conference on Learning Representations}.

\bibitem[{Wijk et~al.(2025)Wijk, Lin, Becker, Jawhar, Parikh, Broadley, Chan, Chen, Clymer, Dhyani, Ericheva, Garcia, Goodrich, Jurkovic, Karnofsky, Kinniment, Lajko, Nix, Sato, Saunders, Taran, West, and Barnes}]{wijk2024rebench}
Hjalmar Wijk, Tao Lin, Joel Becker, Sami Jawhar, Neev Parikh, Thomas Broadley, Lawrence Chan, Michael Chen, Josh Clymer, Jai Dhyani, Elena Ericheva, Katharyn Garcia, Brian Goodrich, Nikola Jurkovic, Holden Karnofsky, Megan Kinniment, Aron Lajko, Seraphina Nix, Lucas Sato, and 4 others. 2025.
\newblock \href {https://arxiv.org/abs/2411.15114} {Re-bench: Evaluating frontier ai {R\&D} capabilities of language model agents against human experts}.
\newblock \emph{Preprint}, arXiv:2411.15114.

\bibitem[{Xu et~al.(2021)Xu, Liu, Cao, Huang, Liu, Qian, Liu, Wu, Dong, Qiu, Qiu, Hua, Su, Wu, Xu, Han, Fu, Yin, Liu, Roepman, Dietmann, Virta, Kengara, Zhang, Zhang, Zhao, Dai, Yang, Lan, Luo, Liu, An, Zhang, He, Cong, Liu, Zhang, Lewis, Tiedje, Wang, An, Wang, Zhang, Huang, Lu, Cai, Wang, and Zhang}]{xu2021powerful}
Yongjun Xu, Xin Liu, Xin Cao, Changping Huang, Enke Liu, Sen Qian, Xingchen Liu, Yanjun Wu, Fengliang Dong, Cheng-Wei Qiu, Junjun Qiu, Keqin Hua, Wentao Su, Jian Wu, Huiyu Xu, Yong Han, Chenguang Fu, Zhigang Yin, Miao Liu, and 29 others. 2021.
\newblock \href {https://doi.org/10.1016/j.xinn.2021.100179} {Artificial intelligence: A powerful paradigm for scientific research}.
\newblock \emph{The Innovation}, 2(4):100179.
\newblock Review article.

\bibitem[{Yamada et~al.(2025)Yamada, Lange, Lu, Hu, Lu, Foerster, Clune, and Ha}]{yamada2025aiscientistv2}
Yutaro Yamada, Robert~Tjarko Lange, Cong Lu, Shengran Hu, Chris Lu, Jakob Foerster, Jeff Clune, and David Ha. 2025.
\newblock \href {https://arxiv.org/abs/2504.08066} {The ai scientist-v2: Workshop-level automated scientific discovery via agentic tree search}.
\newblock \emph{Preprint}, arXiv:2504.08066.

\bibitem[{Yu et~al.(2025)Yu, Hong, Cheng, Zhu, Xuan, Yao, Feng, and You}]{yu2024researchtown}
Haofei Yu, Zhaochen Hong, Zirui Cheng, Kunlun Zhu, Keyang Xuan, Jinwei Yao, Tao Feng, and Jiaxuan You. 2025.
\newblock \href {https://arxiv.org/abs/2412.17767} {Researchtown: Simulator of human research community}.
\newblock \emph{Preprint}, arXiv:2412.17767.

\bibitem[{Zheng et~al.(2023)Zheng, Koh, Ju, Nguyen, May, Webb, and Pan}]{zheng2023large}
Yizhen Zheng, Huan~Yee Koh, Jiaxin Ju, Anh T.~N. Nguyen, Lauren~T. May, Geoffrey~I. Webb, and Shirui Pan. 2023.
\newblock \href {https://arxiv.org/abs/2310.07984} {Large language models for scientific synthesis, inference and explanation}.
\newblock \emph{Preprint}, arXiv:2310.07984.

\end{thebibliography}

\clearpage
\onecolumn
\appendix
\section{LLMs' Discussion Prompts for Research Ideation}
\label{sec:discussion_prompts}
\noindent This section reproduces the exact system prompts used for the ideation–critique–revision pipeline, ensuring full reproducibility of our multi-agent dialogue experiments. We adopt the prompt design originally proposed by~\citet{si2025can}.

{\centering
\begin{tcolorbox}[
    colframe=black!50!white, 
    colback=black!10!white, 
    title=Initial Idea Generation Prompt, 
    width=0.95\textwidth,
    boxsep=3pt
]
\footnotesize
\begin{Verbatim}[breaklines, breakanywhere=true, obeytabs=true]
{persona_prompts}

Now I want you to help me brainstorm some new research project ideas on the topic of: {topic_description}.

Here are some relevant papers on this topic just for your background knowledge: {formatted_papers}

You should generate {ideas_n} different ideas on this topic. Try to be creative and diverse in the idea generation, and do not repeat any similar ideas. The above papers are only for inspiration and you should not cite them and just make some incremental modifications. Instead, you should make sure your ideas are novel and distinct from the prior literature. You should aim for projects that can potentially win best paper awards at top AI conferences like ACL and NeurIPS.

Each idea should be described as:
  (1) Problem: State the problem statement, which should be closely related to the topic description and something that large language models cannot solve well yet.
  (2) Existing Methods: Mention some existing benchmarks and baseline methods if there are any.
  (3) Motivation: Explain the inspiration of the proposed method and why it would work well.
  (4) Proposed Method: Propose your new method and describe it in detail. The proposed method should be maximally different from all existing work and baselines, and be more advanced and effective than the baselines. You should be as creative as possible in proposing new methods, we love unhinged ideas that sound crazy. This should be the most detailed section of the proposal.
  (5) Experiment Plan: Specify the experiment steps, baselines, and evaluation metrics.

You can follow these examples to get a sense of how the ideas should be formatted (but don't borrow the ideas themselves): {examples}

You should make sure to come up with your own novel and different ideas for the specified problem: {topic_description}. 

You should try to tackle important problems that are well recognized in the field and considered challenging for current models. For example, think of novel solutions for problems with existing benchmarks and baselines. In rare cases, you can propose to tackle a new problem, but you will have to justify why it is important and how to set up proper evaluation. 

Focus on novel {method} ideas for now. The proposed method section should specify how to construct the prompts for all steps involved. Try to avoid large-scale pretraining experiments or human studies.

You should avoid repeating the following existing ideas and try to be different and diverse: {existing_ideas}

Please write down your {ideas_n} ideas (each idea should be described as one paragraph). 
Output the ideas in json format as a dictionary, where you should generate a short idea name (e.g., "Non-Linear Story Understanding", or "Multi-Agent Negotiation") as the key and the actual idea description as the value (following the above format). Do not repeat idea names or contents.
\end{Verbatim}
\end{tcolorbox}
\par}
\newpage

{\centering
\begin{tcolorbox}[
    colframe=black!50!white, 
    colback=black!10!white, 
    title=Critique Prompt, 
    width=0.95\textwidth,
    boxsep=3pt
]
\footnotesize
\begin{Verbatim}[breaklines, breakanywhere=true, obeytabs=true]
{persona_prompts}

You need to provide some constructive feedback to the given project proposal on the topic of: {topic_description}.

The project proposal (containing multiple ideas) is

{current_ideas_json_str}

You should raise critical questions and comments if there are any missing details from the project proposal, or if there are any parts that are not feasible for the student to complete the project within two months. 

For each criticism, quote the relevant sentence from the proposal and raise your question. Your criticisms should be a subset of the following categories:
- missing dataset detail: the proposal should mention the datasets to use, or how to collect the data if needed
- involving humans: try to avoid human experiments
- missing metric detail: the proposal should mention the evaluation metrics
- missing prompt detail: give concrete examples of the prompts (vague descriptions are not enough), including for the new proposed prompting method and for all baselines
- test cases: the proposal should show 1-2 examples of how the test examples would look like, how the prompts will be applied, and the expected outputs

Each criticism should quote a sentence from the original proposal, and try to be concise. Here are a few example criticisms:

Missing data preparation detail: "Create a comprehensive list of idiomatic expressions" - how to create such a list? Any tools or sources that we can use?

Involving human experiments: "Contrast the LLM's responses with a baseline of human expectations" - we should avoid human annotation to make the project more feasible, try to think of alternatives with automatic approaches, such as using LLMs to replace humans.

Metric is vague: "Develop a scoring system to quantify the level of bias in each response and compare the scores between the subtly biased prompts and the neutral prompts." - how exactly should we implement such a scoring system? There should be detailed step-by-step instructions.

Prompt not specified: "Develop a set of prompts that describe scenarios involving individuals in both stereotypical and counter-stereotypical roles." - you should always try to give the exact prompts to use.

Missing test cases: "Test the model on a set of prompts and compare the responses to human expectations" - add a subsection in the experiment plan to show how an example prompt is applied on a test example and what the expected output is.

Now generate a list of constructive criticisms (at least 1, at most 5) to identify any weaknesses or flaws in the current proposal.
\end{Verbatim}
\end{tcolorbox}
\par}
\newpage

{\centering
\begin{tcolorbox}[
    colframe=black!50!white, 
    colback=black!10!white, 
    title=Revise Prompt, 
    width=0.95\textwidth,
    boxsep=3pt
]
\footnotesize
\begin{Verbatim}[breaklines, breakanywhere=true, obeytabs=true]
{persona_prompts}

You previously proposed the following research ideas on the topic of: {topic_description}.

The original project proposal (containing multiple ideas) is

{current_ideas_json_str}

However, the following criticisms were raised by expert reviewers: {response_critic}

Please revise the original research proposal based on the criticisms provided above. Your goal is to improve the ideas by addressing each point raised in the critique, making the proposal more concrete, feasible, novel, and impactful.

Specifically, you should:
- Carefully consider each criticism and modify the corresponding idea(s) to resolve the identified issues.
- Ensure all necessary details are included, such as specific datasets, evaluation metrics, concrete prompt examples, and clear experimental steps, as pointed out by the critique.
- If an idea is fundamentally flawed or infeasible according to the critique, you may replace it with a new, improved idea on the same topic, ensuring it still aligns with the overall research goal.
- Maintain the original structure for each idea: (1) Problem, (2) Existing Methods, (3) Motivation, (4) Proposed Method, (5) Experiment Plan.

Output the revised set of ideas in the original JSON format: a dictionary where keys are short, descriptive idea names and values are the detailed idea descriptions following the specified structure. Ensure you provide the same number of ideas as the original proposal.
\end{Verbatim}
\end{tcolorbox}
\par}
\newpage

\section{Diverse Persona Prompts}
\label{sec:persona_prompts}
This section provides the domain-specific persona instructions that inject disciplinary perspectives into individual agents during the dialogue.

{\centering
\begin{tcolorbox}[
    colframe=blue!50!black!40!white, 
    colback=blue!10!white, 
    title=Baseline Persona Prompt, 
    width=0.95\textwidth,
    boxsep=3pt
]
\footnotesize
\begin{Verbatim}[breaklines, breakanywhere=true, obeytabs=true]
You are an expert AI researcher.
\end{Verbatim}
\end{tcolorbox}
\par}

\vspace{1em}

{\centering
\begin{tcolorbox}[
    colframe=blue!50!black, 
    colback=blue!10!white, 
    title=``Physics-AI'' Persona Prompt, 
    width=0.95\textwidth,
    boxsep=3pt
]
\footnotesize
\begin{Verbatim}[breaklines, breakanywhere=true, obeytabs=true]
You are an expert in physics and its intersection with AI. Propose novel AI research methodologies (e.g., algorithms, models) inspired by physical principles (like statistical mechanics, field theory, or complex systems dynamics), or critique existing AI approaches using insights and analogies from physics.
\end{Verbatim}
\end{tcolorbox}
\par}

{\centering
\begin{tcolorbox}[
    colframe=blue!50!black, 
    colback=blue!10!white, 
    title=``Chemistry-AI'' Persona Prompt, 
    width=0.95\textwidth,
    boxsep=3pt
]
\footnotesize
\begin{Verbatim}[breaklines, breakanywhere=true, obeytabs=true]
You are an expert in chemistry and its intersection with AI. Propose novel AI research methodologies (e.g., learning processes, representations) inspired by chemical concepts (like reaction kinetics, molecular interactions, or self-assembly), or critique existing AI approaches using insights and analogies from chemistry.
\end{Verbatim}
\end{tcolorbox}
\par}

{\centering
\begin{tcolorbox}[
    colframe=blue!50!black, 
    colback=blue!10!white, 
    title=``Biology-AI'' Persona Prompt, 
    width=0.95\textwidth,
    boxsep=3pt
]
\footnotesize
\begin{Verbatim}[breaklines, breakanywhere=true, obeytabs=true]
You are an expert in biology and its intersection with AI. Propose novel AI research methodologies (e.g., architectures, learning rules) inspired by biological systems (like evolution, neuroscience, or immune systems), or critique existing AI approaches using insights and analogies from biology.
\end{Verbatim}
\end{tcolorbox}
\par}

{\centering
\begin{tcolorbox}[
    colframe=blue!50!black, 
    colback=blue!10!white, 
    title=``Finance-AI'' Persona Prompt, 
    width=0.95\textwidth,
    boxsep=3pt
]
\footnotesize
\begin{Verbatim}[breaklines, breakanywhere=true, obeytabs=true]
You are an expert in finance and its intersection with AI. Propose novel AI research methodologies (e.g., decision-making models, predictive algorithms) inspired by financial concepts (like market dynamics, behavioral economics, or risk modeling), or critique existing AI approaches using insights and analogies from finance.
\end{Verbatim}
\end{tcolorbox}
\par}

{\centering
\begin{tcolorbox}[
    colframe=blue!50!black, 
    colback=blue!10!white, 
    title=``Psychology-AI'' Persona Prompt, 
    width=0.95\textwidth,
    boxsep=3pt
]
\footnotesize
\begin{Verbatim}[breaklines, breakanywhere=true, obeytabs=true]
You are an expert in psychology and its intersection with AI. Propose novel AI research methodologies (e.g., reasoning processes, learning paradigms, interaction designs) inspired by psychological theories (like cognitive processes, learning theories, or social dynamics), or critique existing AI approaches using insights and analogies from psychology.
\end{Verbatim}
\end{tcolorbox}
\par}

\clearpage

\section{Examples of Generated Research Ideas}
This section showcases representative outputs—generated for the topic ``novel prompting methods to reduce social biases and stereotypes of large language models''—under each dialogue configuration to illustrate qualitative differences in the generated ideas.

\subsection{No Critique (Single)}
{\centering
\begin{tcolorbox}[
    colframe=green!50!black, 
    colback=green!10!white, 
    title=Contextual Anchoring: A Novel Prompting Method to Mitigate Bias in Large Language Models, 
    width=0.95\textwidth,
    boxsep=3pt
]
\# Problem Statement \\
Large language models (LLMs) often reflect and amplify existing biases present in their training data, leading to biased outputs that reinforce stereotypes. This is particularly problematic when prompts lack sufficient grounding in diverse contexts, resulting in oversimplified or skewed representations of social issues.
\\
\\
\# Motivation \\
Existing methods, such as few-shot prompting and debiasing prompts, have shown limited effectiveness due to their reliance on static examples that may not capture the complexity of human diversity. Our proposed method, inspired by contextual anchoring, aims to mitigate bias by constructing prompts that include demographic, cultural, and situational markers. By encouraging LLMs to consider multiple perspectives and societal contexts simultaneously, we hypothesize that this method will yield more nuanced and less biased outputs compared to traditional prompting methods.
\\
\\
\# Proposed Method \\
The proposed method involves the following steps: \\
1) Identify key demographic, cultural, and situational markers relevant to the query. \\
2) Construct prompts that integrate these markers with the core question. For example, instead of asking, 'What is the role of women in tech?', the prompt would be rephrased to, 'In the context of diverse global societies, what are some roles women occupy in technology fields, particularly focusing on Western and Eastern perspectives?'. \\
3) Use these prompts to query LLMs and analyze the outputs for bias and complexity.\\
\end{tcolorbox}
\par}
\clearpage

\subsection{Baseline (Self-Critique)}
{\centering
\begin{tcolorbox}[
    colframe=green!50!black, 
    colback=green!10!white, 
    title=Contextual Bridging Prompting: Enhancing Code Generation in Modular Codebases with Large Language Models, 
    width=0.95\textwidth,
    boxsep=3pt
]
\# Problem Statement \\
Current code generation models often struggle to maintain context when generating multi-file or modular codebases, leading to inconsistencies and failed integrations between modules. This issue is particularly critical in software development, where modularity and interdependencies are essential for functionality and maintainability.
\\
\\
\# Motivation \\
Existing methods primarily rely on basic sequential prompting, which fails to capture the relationships between code modules. This results in isolated code snippets that do not consider interdependencies, leading to integration issues. Our proposed method, Contextual Bridging Prompting, draws inspiration from human cognitive processes in software design, where developers outline module interactions and dependencies before implementation. By structuring the generation process around these interdependencies, we aim to improve the coherence and functionality of generated code.
\\
\\
\# Proposed Method \\
The Contextual Bridging Prompting method consists of three main steps: \\
1) Identify Functional Requirements: Prompt the LLM to outline the broader functional requirements and necessary interfacing between modules. \\
2) Create a Blueprint: Ask the LLM to generate a blueprint or architecture that illustrates the relationships between modules. \\
3) Generate Module Code: Craft prompts for individual modules that reference the established blueprint, ensuring that each module's implementation considers its interactions with other modules.
\end{tcolorbox}
\par}
\clearpage

\subsection{Parallel Self-Critique (N = 4)}
{\centering
\begin{tcolorbox}[
    colframe=green!50!black, 
    colback=green!10!white, 
    title=Argumentation-Based Prompting for Enhanced Code Generation in Large Language Models, 
    width=0.95\textwidth,
    boxsep=3pt
]
\# Problem Statement \\
Large Language Models (LLMs) often struggle with structured reasoning, particularly when faced with trade-offs between competing code solutions. This limitation can lead to suboptimal code outputs, especially in scenarios requiring nuanced decision-making.
\\
\\
\# Motivation \\
Existing prompting strategies, including standard and chain-of-thought methods, do not effectively encourage LLMs to explore alternative solutions or articulate justifications for their choices. This lack of depth in reasoning can result in less optimal code generation. By drawing on formal argumentation principles, we propose a method that encourages LLMs to generate multiple code solutions and evaluate their pros and cons, thereby enhancing the reasoning process and improving code quality.
\\
\\
\# Proposed Method \\
The proposed method, Argumentation-Based Prompting, involves a two-step process: \\
(1) generating multiple code solutions for a given requirement, and \\
(2) evaluating these solutions by articulating the pros and cons of each. The prompts will guide the model to reason about trade-offs, such as performance, readability, and dependencies. For example, a prompt could be: 'Generate three different implementations for the sorting requirement and argue why Solution A may be preferable due to fewer dependencies and better performance.    
\end{tcolorbox}
\par}
\clearpage

\subsection{Iterative Self-Critique (L = 4)}
{\centering
\begin{tcolorbox}[
    colframe=green!50!black, 
    colback=green!10!white, 
    title=Interactive Context Shaping: Enhancing Code Generation in Large Language Models through Collaborative Prompting, 
    width=0.95\textwidth,
    boxsep=3pt
]
\# Problem Statement \\
Large language models (LLMs) often struggle with contextually nuanced code generation tasks, particularly when multiple requirements need to be synthesized for coherent output. This leads to suboptimal code generation that may not meet user expectations or requirements.
\\
\\
\# Motivation \\
Existing methods, such as chain-of-thought (CoT) prompting, rely on sequential reasoning but can lose track of context when faced with dynamic requirements or multiple objectives. This project draws inspiration from human collaborative problem-solving, where individuals refine and reshape context through interactive dialogues. By simulating a collaborative environment, we hypothesize that LLMs can produce more coherent and contextually aligned code outputs, leading to improved performance in code generation tasks.
\\
\\
\# Proposed Method \\
We introduce Interactive Context Shaping (ICS), a multi-step prompting process that includes: \\
1) Initial Requirement Gathering: Prompt the LLM to list precise requirements needed for code generation based on a vague requirement. \\
2) Contextual Shaping: Organize these requirements into distinct categories, highlighting dependencies and sequential steps. \\
3) Iterative Refinement: After generating code, prompt the LLM to review the code against the initial requirements and outline any discrepancies for further refinement. This interactive loop ensures continual refinement and deeper understanding of each requirement.",  
\end{tcolorbox}
\par}
\clearpage

\subsection{Diverse Personas (Critic)}
{\centering
\begin{tcolorbox}[
    colframe=green!50!black, 
    colback=green!10!white, 
    title=Goal-Oriented Prompting with Iterative Refinement for Enhanced Code Generation in Large Language Models, 
    width=0.95\textwidth,
    boxsep=3pt
]
\# Problem Statement \\
Large Language Models (LLMs) often generate code that meets initial requirements but fails to optimize for best practices or efficiency. This leads to suboptimal code that may not adhere to performance metrics such as runtime and memory usage.
\\
\\
\# Motivation \\
Existing methods, including simple iterative feedback loops, do not focus on specific goals, resulting in inadequate code comprehension and enhancement. By integrating goal-oriented framing within the prompting process, we can better align code output toward optimized performance. The proposed Goal-Oriented Prompting with Iterative Refinement (GOPIR) method aims to guide LLMs through a structured iterative process that emphasizes specific performance metrics, thereby improving the quality of generated code.
\\
\\
\# Proposed Method \\
The GOPIR method consists of the following steps: \\
1) Generate initial code based on a requirement prompt. \\
2) Evaluate the generated code using goal-oriented prompts that focus on specific performance metrics (e.g., efficiency, clarity). \\
3) Suggest improvements based on the evaluation. \\
4) Refine the code based on the suggested improvements. \\
5) Repeat the evaluation and refinement process for a specified number of iterations or until no further improvements are suggested.
\end{tcolorbox}
\par}
\clearpage

\subsection{Diverse Personas (Prop/Rev)}
{\centering
\begin{tcolorbox}[
    colframe=green!50!black, 
    colback=green!10!white, 
    title=Hierarchical Specification Prompting: Enhancing Code Generation in Large Language Models, 
    width=0.95\textwidth,
    boxsep=3pt
]
\# Problem Statement \\
Current large language models (LLMs) struggle to generate complex code that adheres to multi-level specifications, often resulting in incomplete or incorrect implementations. This issue is particularly evident in benchmarks like HumanEval and MBPP, where models perform poorly on tasks requiring meticulous adherence to detailed requirements.
\\
\\
\# Motivation \\
Existing prompting techniques, such as few-shot and chain-of-thought prompting, typically present a single flat requirement, failing to leverage the multi-level nature of specifications. This limitation leads to performance drops on complex tasks. Our proposed method, Hierarchical Specification Prompting (HSP), aims to structure prompts to reflect multiple layers of expectations, guiding LLMs to generate code that comprehensively meets all requirements. By framing prompts in a hierarchical manner, we can better align the model's output with the detailed specifications.
\\
\\
\# Proposed Method \\
The proposed method involves three sequential prompting steps: \\
(1) **High-Level Overview**: Prompt the model to develop a high-level overview of the feature based on a given feature description. \\
- Example Prompt: 'Please develop a high-level overview of the feature: [Feature Description]' \\
(2) **Functional Requirements**: Prompt the model to outline specific components needed, including their functionalities based on the high-level overview. \\
- Example Prompt: 'Outline the specific components needed, including their functionalities: [Functional Requirements]' \\
(3) **Code Generation**: Prompt the model to generate code for each component based on detailed specifications. \\
- Example Prompt: 'Generate code for each component based on specifications, e.g., [Component Specifications].' \\
Each segment builds on the outputs of the previous segments to ensure a thorough understanding before code generation.",
                
\end{tcolorbox}
\par}
\clearpage

\end{document}